\documentclass[manuscript]{acmart}
\usepackage{float}
\usepackage{amsmath}
\newtheorem{myDef}{Definition}
\usepackage{graphicx}  
\usepackage{bm}
\usepackage{algorithm, algorithmic}
\usepackage{caption}
\usepackage{subfigure}
\usepackage{makecell}
\usepackage{wrapfig}
\usepackage{stfloats}
\usepackage{bbm}
\usepackage{multirow}
\usepackage{makecell}
\usepackage{colortbl}
\hypersetup{draft}
\definecolor{mygray}{gray}{.9}
\newtheorem{thm}{\bf Proposition}[section]

\AtBeginDocument{%
  \providecommand\BibTeX{{%
    \normalfont B\kern-0.5em{\scshape i\kern-0.25em b}\kern-0.8em\TeX}}}

\setcopyright{acmcopyright}
\copyrightyear{2018}
\acmYear{2018}
\acmDOI{10.1145/1122445.1122456}

\acmConference[Woodstock '18]{Woodstock '18: ACM Symposium on Neural
  Gaze Detection}{June 03--05, 2018}{Woodstock, NY}
\acmBooktitle{Woodstock '18: ACM Symposium on Neural Gaze Detection,
  June 03--05, 2018, Woodstock, NY}
\acmPrice{15.00}
\acmISBN{978-1-4503-XXXX-X/18/06}

\begin{document}

\title{Revisiting the role of heterophily in graph representation learning: An edge classification perspective}

\author{Jincheng Huang}
\affiliation{%
  \institution{School of Computer Science, Southwest Petroleum University}
 \city{Chengdu}
  \country{China}
  }
 \email{huangjc0429@gmail.com}

 \author{Ping Li}
 \authornote{Corresponing Author.}
 \affiliation{%
  \institution{School of Computer Science, Southwest Petroleum University}
   \city{Chengdu}
  \country{China}}
 \email{dping.li@gmail.com}

 \author{Rui Huang}
 \affiliation{%
   \institution{School of Computer Science, Southwest Petroleum University}
  \city{Chengdu}
  \country{China}}
 \email{huangrui1104@foxmail.com}

 \author{Na Chen}
 \affiliation{%
 \institution{School of Computer Science, Southwest Petroleum University}
  \city{Chengdu}
  \country{China}}
 \email{dpnanachen@gmail.com}
 
  \author{Acong Zhang}
 \affiliation{%
 \institution{School of Computer Science, Southwest Petroleum University}
  \city{Chengdu}
  \country{China}}
 \email{zac1328682511@gmail.com}

\begin{abstract}
Graph representation learning aim at integrating node contents with graph structure to learn nodes/graph representations. Nevertheless, it is found that many existing graph learning methods do not work well on data with high heterophily level that accounts for a large proportion of edges between different class labels. Recent efforts to this problem focus on improving the message passing mechanism. However, it remains unclear whether heterophily truly does harm to the performance of graph neural networks (GNNs). The key is to unfold the relationship between a node and its immediate neighbors, 
e.g., are they heterophilous or homophilious? From this perspective, here we study the role of heterophily in graph representation learning before/after the relationships between connected nodes are disclosed. In particular, we propose an end-to-end framework that both learns the type of edges (i.e., heterophilous/homophilious) and leverage edge type information to improve the expressiveness of graph neural networks.
We implement this framework in two different ways. Specifically, to avoid messages passing through heterophilous edges, we can optimize the graph structure to be homophilious by dropping heterophilous edges identified by an edge classifier. Alternatively, it is possible to exploit the information about the presence of heterophilous neighbors for feature learning, so a hybrid message passing approach is devised to aggregate homophilious neighbors and diversify heterophilous neighbors based on edge classification. Extensive experiments demonstrate the remarkable performance improvement of GNNs with the proposed framework on multiple datasets across the full spectrum of homophily level.

\end{abstract}



\maketitle

\section{Introduction}
\noindent The surprising expressiveness of graph neural networks (GNNs) has aroused an explosion of interests in graph representation learning~\cite{hamilton2020graph}, resulting in extensive applications ranging from social network analysis~\cite{socialnetwork} to molecular biology~\cite{molecular1,molecular2}, even to regular data processing like text mining~\cite{nlp} and image processing~\cite{cv1}. Despite their practical success, it is not yet guaranteed that GNNs can be effective for arbitrary graph data.

\begin{figure}[t]
	\centering
	\includegraphics[scale=0.3]{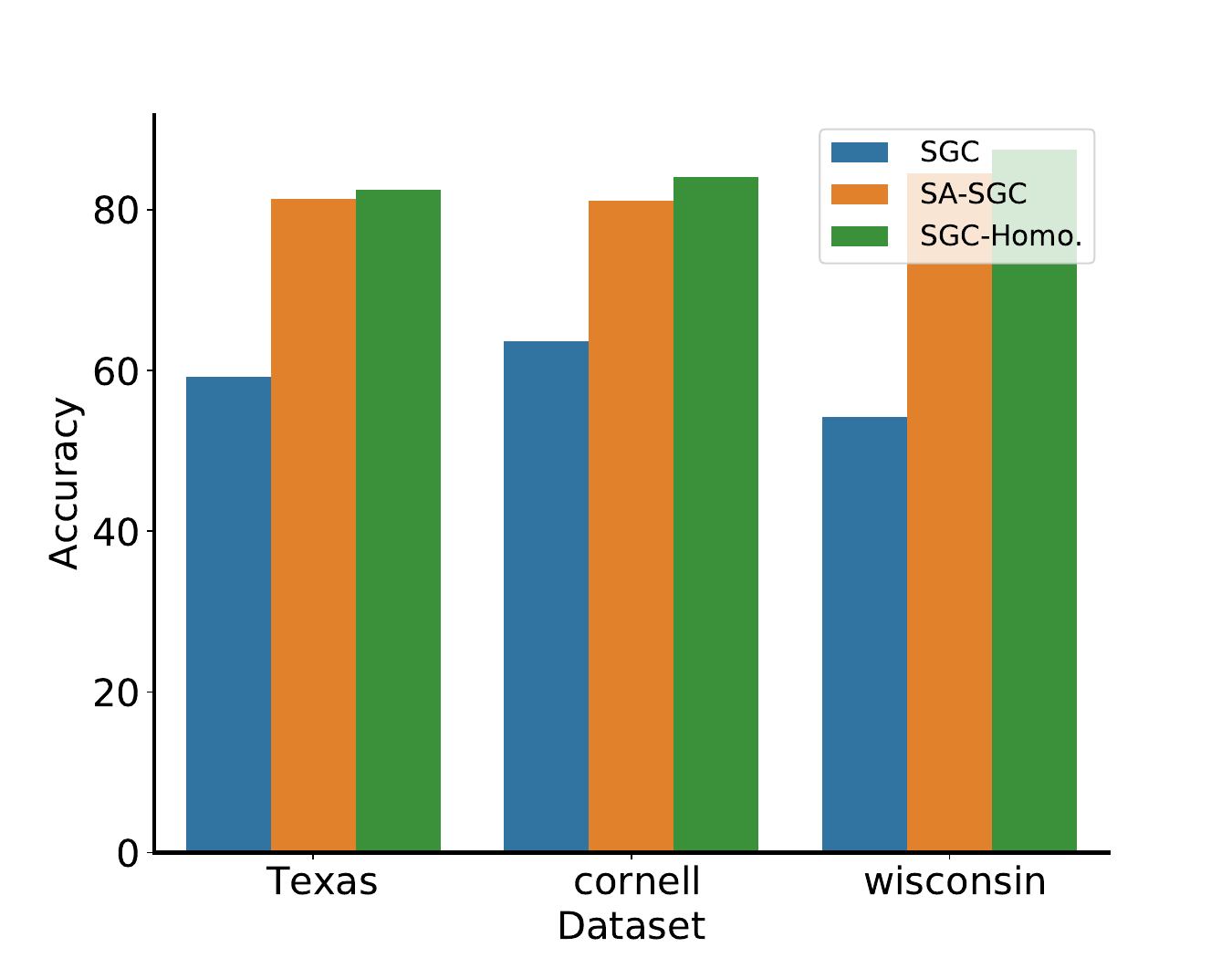}
	\caption{Performance of SGC on graphs in different settings: the blue column corresponds to the vanilla SGC on the original datasets, orange column refers to SGC equipped with the proposed framework, while green column reflects the performance of the vanilla SGC on perfectly homophilious counterparts of the original graphs.}
	\label{fig:bound}
\end{figure}

One notable feature in graphs is heterophily~\cite{zhu2020}: connected nodes may have different class labels or properties. In real-world systems, many graphs exhibit strong heterophily. For instance, the matchmaking website tends to connect people to those with the opposite gender, the hierarchical organization structure of many natural and man-made systems shows the leader-member relations. When 
applying GNNs on such graph data, the implicit assumption underlying GNNs that node features should be similar (smooth) among neighboring nodes is violated, which makes it very possible to learn undiscriminating  features for different classes. 

To tackle this issue, a few recent work has made efforts to capture node feature encoded in the topology from different respects. For example, by looking into the distribution of class labels in a graph, Zhu et al.~\cite{zhu2020} find that the 2-hop neighborhood is less heterophilious than the nearest neighborhood, thus it is expected to improve the power of graph learning by leveraging the second-order neighbors. Geom-GCN~\cite{pei2020} maps a graph to a continuous space to capture long-range dependencies between nodes such that the information on distant nodes with the same label is able to be exploited in node representation learning. On the other hand, it is possible to aggregate messages from different neighbors (i.e., homophilious neighbors and heterophilous neighbors) in different ways based on the semantic similarity between neighboring nodes~\cite{gbkgnn, ACM}. Though the prior work has carefully crafted the way of information integration, to optimize representation learning strategy needs to exhaustively examine the characteristics of class label distribution in a graph, which remains a great challenge.

\begin{figure*}[h]
	\centering
	\includegraphics[scale=0.5]{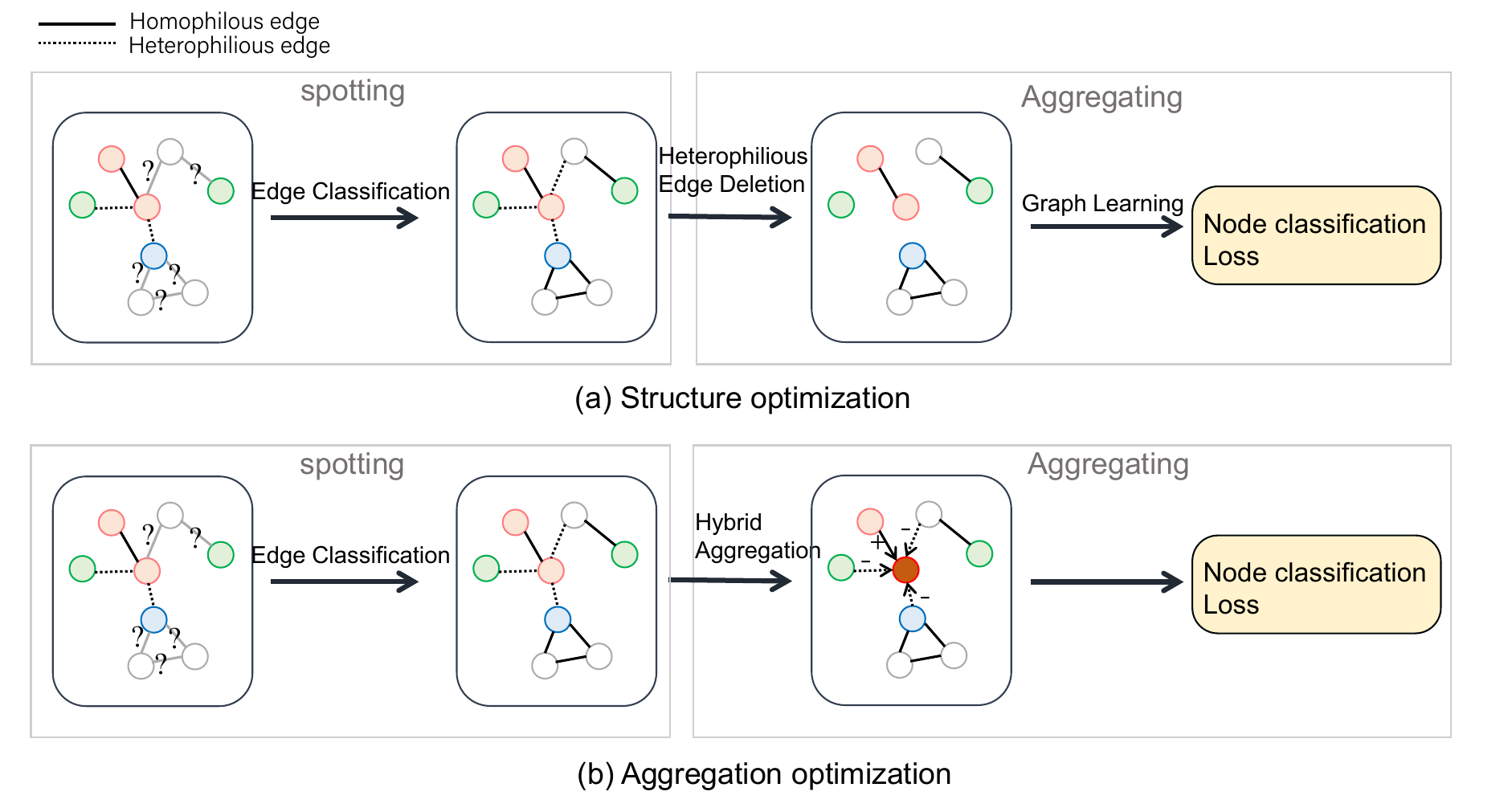}
	\caption{Two implementations of our ''spotting-then-aggregating'' framework: (a) Structure optimization based on edge type spotting. (b) Aggregation optimization based on edge type spotting.}
	\label{fig:framework}
\end{figure*}

Clearly the block is the presence of heterophilious edges. As shown in Figure~\ref{fig:bound}, even by simply removing all heterophilous edges in graph data, the performance of a state-of-the-art model SGC~\cite{2019sgc} can be improved by about 30 percentage points, compared to the original data. This result suggests that under the classical message passing scheme,  heterophilous edges are indeed detrimental to representation learning. 
Therefore, an intuitive solution is to use only homophilious edges for message passing, which however requires that edge type is available. Furthermore, once edge type is available, considering that the presence of heterophilous edges can be seen as a feature and play a role in bridging long distance nodes, it will be beneficial to exploit heterophilous edges in feature representation. In effect, previous study~\cite{gbkgnn} shows that even a rough estimate on edge type by attention mechanism facilitates the mixing of incompatible features. However, different from existing feature based attention models ~\cite{gbkgnn}, the use of edge type makes it more direct and efficient to learn appropriate feature aggregation on the neighborhood. Thus, it is meaningful to identify edge types in graphs.

 In this work, we propose an end-to-end ''spotting-then-aggregating'' scheme for graph representation learning, which consists of two components: edge type identification (i.e., ''spotting'') and node feature representation learning (i.e., ''aggregating''). Specifically, we categorize the edges into two types, namely, homophilious edge and heterophilious edge, where homophilious edges refer to the edges connecting nodes in the same class, and heterophilious edges represent the edges connecting nodes between different classes. Based on the edge labeling on the training set, a binary classification model is learned to discriminate between heterophilious and homophilious edges, with which the heterophilious edges are spotted from the unlabeled set. On top of that, representation learning can be improved. Here we provide two strategies for improving graph representation learning, namely, removing detected heterophilous edges from the graph, or aggregating messages using two channels (corresponding to homophilious edges and heterophilous edges respectively). 
 
 The whole process is implemented in an end-to-end manner, which allows the edge type identification to be dynamically adjusted according to the prediction of GNNs. Despite its adaptability, end-to-end learning will suffer from discrete meta-outputs. In particular, the output of edge classifier is binary values, which means that continuous gradients cannot be propagated back to the edge classifier. To resolve this problem, we devise a back-max method based on Gumbel-softmax~\cite{gumbelsoftmax2017} to tune the parameters of the edge classifier by the task-specific loss. 
 
 The contribution of this paper is summarized as follows:

	\begin{itemize}
	\item We present a perspective from edge type identification to study the role of heterophily in graph representation learning, which makes it possible to leverage the presence of heterophily to boost graph neural networks.
	
	\item We propose the general "spotting-then-aggregating" framework
	to learn the heterophily of the edges and explore the use of heterophilious edges. In particular, we devise two simple yet effective methods to implement the framework.
	
	\item  Extensive experiments on the benchmark graph data with low homophily validate the superiority of the proposed method over the state-of-the-art models.
\end{itemize}

\section{Preliminaries}
\label{sec:prelimit}
Let \bm{$G = (V,E)$} be an undirected and unweighted graph with node set $V$ and edge set $E$. The nodes are described by the feature matrix~\bm{ $X \in\mathbb{R}^{n\times f}$}, where \bm{$f$} denotes the number of features per node and $n$ is the number of nodes. Each node is associated with a class label, which is depicted in the label matrix \bm{$Y \in\mathbb{R}^{n\times c}$} with a total of \bm{$c$} classes. We represent the graph by its adjacency matrix \bm{$A \in \mathbb{R}^{n\times n}$} and the graph with self-loops by \bm{$\tilde{A} = A + I_{n}$}.\\

\noindent\textbf{Graph Convolutional Network and Its Simplification} Message passing graph convolutional networks have been proved to be powerful on a number of graph data, among which graph convolutional network(GCN) proposed by Kipf et al.~\cite{kipf2017gcn} is a widely used one. A typical GCN makes prediction according to a series of operations on node features:
\begin{equation}\label{eq-1}
	\bm{\widehat{Y}} = softmax(S_{sym}ReLU(S_{sym}XW_{0})W_{1}),
\end{equation}
where \bm{$\widehat{Y}\in \mathbb{R}^{n\times c}$} are the predicted node labels, {\bm$W_{0}$} and {\bm$W_{1}$} are for feature mapping at the first and second layer, respectively. Besides, \bm{$S_{sym} = \tilde{D}^{-\frac{1}{2}}\tilde{A}\tilde{D}^{-\frac{1}{2}}$} denotes the symmetrically normalized adjacency matrix with self-loops, where \bm{$\tilde{D}$} refers to the diagonal degree matrix of \bm{$\tilde{A}$}.

The above Equation~\ref{eq-1} reveals three core components in GCN, namely, feature propagation, linear transformation, non-linear activation. However, Wu et al.~\cite{2019sgc} have shown that the latter two components are redundant and detrimental to performance, so they simplify the vanilla GCN as

\begin{equation}
	\bm{\widehat{Y}} = softmax(S_{sym}^{K}XW),
\end{equation}
where $K$ is the number of graph convolutions, and $W\in\mathbb{R}^{d\times c}$ denotes the learnable parameters of a logistic regression classifier. The simplified GCN is referred to as SGC.\\

\textbf{Graph Fourier Transform and Graph Signals.} In the light of graph signal theory~\cite{graph_fourier}, graph convolution is equivalent to the Laplacian transform of graph signals from the time domain to the frequency domain. Let $L = I_{n} - D^{-\frac{1}{2}}AD^{-\frac{1}{2}}$ be the normalized graph Laplacian matrix, which is positive semi-definite and has a complete set of orthogonal eigenvectors $\{u_{l}\}^{n}_{l=1} \in \mathbb{R}^{n}$ corresponding to the eigenvalues $\lambda_{l} \in [0, 2]$. Similar to the Laplacian operator to the basis function $e^{-i\omega t}$ of Fourier transform in the time domain, i.e., $\Delta e^{-i\omega t} = \frac{\partial ^{2}e^{-i\omega t}}{\partial t^{2}} = - \omega ^{2}e^{-i\omega t}$, the eigenvectors of Laplacian matrix is analogous to the basis functions of Fourier transform. That is, the Fourier transform on the graph can be defined as $\hat{x} = U^{T}x$, associated with its inverse transform $x = U\hat{x}$. Hence, the convolution of graph signal $x$ with the convolution kernel $f$ reads as:
\begin{equation}
\label{eq:graph_four}
    (f\ast g)_{G} = U((U^{T}f)\odot  (U^{T}x)) = Ug_{\theta}U^{T}x,
\end{equation}
where $\odot$ is hadamard product, and $g_{\theta}$ is a diagonal matrix, which represents the convolutional kernel in the frequency domain, replacing $U^{T}f$. 

\textbf{Homophily and heterophily} In this work, we use the homophily measure given in ~\cite{zhu2020} to quantify the homophily level of a graph data, as described in DEFINITION 2. Moreover, we define homophilious and heterophilous edge as :

\begin{myDef}
	Homophilious edge: Edge connecting nodes that have the same label. On the contrary, heterophilious edges are the edges that connect nodes with different labels. We say a graph is homophilious if there are no heterophilious edges in the graph.
	
\begin{myDef}~\cite{zhu2020}
	 The edge homophily ratio is defined as  $h=\frac{|\{(u,v):(u,v)\in\xi \wedge y_{u}=y_{v}\}|}{|\xi|}$, where $|\xi|$  denotes the total number of edges in the graph.
\end{myDef}

\end{myDef}

\section{The impact of heterophily}
\label{sec:pilot}
\begin{table*}[htbp]
	\centering
	\caption{Summary of node classification results (in percent). The * represents the perfectly homophilious situation, and $l$ represents the number of convolution layer. SGC{\scriptsize 2} uses a two-layer MLPs with linear map as the predictor.}
	\label{table:pilot}
	\begin{tabular}{lllllllll}
		\toprule
		Dataset & Cora & Cite. & Film. & Texas & Wisc. & Corn. & Cham. & Squi. \\
		\textit{Homo.ratio} $h$ & 0.81 & 0.74 & 0.22 & 0.06 & 0.1 & 0.2 & 0.23 & 0.22 \\
		\midrule
		\ GCN & 87.2 & 76.4 & 30.1 & 59.5 & 59.8 & 57.3 & 60.3 &  36.7\\
		\rowcolor{mygray}
		\ GCN* & 95.1 & 84.1 & 51.4 & 82.7 & 87.5 & 84.1 & 86.6 & 81.9\\
		\midrule
		\ GAT & 87.6 & 76.3 & 29.0 & 59.1 & 53.1 & 58.4 & 45.1 &  28.3\\
		\rowcolor{mygray}
		\ GAT* & 95.4 & 84.0 & 53.6 & 84.9 & 88.4 & 83.8 & 87.0 &  82.5\\
		\midrule
		\ SGC{\scriptsize 2} & 86.9 & 76.3 & 26.3 & 59.2 & 54.2 & 63.7 & 61.7 & 42.7\\
		\rowcolor{mygray}
		\ SGC{\scriptsize 2}*(l=2)  & 95.6 & 84.2 & 59.4 & 83.5 & 87.7 & 83.0 & 87.2 & 81.9\\
		\rowcolor{mygray}
		\ SGC{\scriptsize 2}*(l=50)  & 96.2 & 84.9 & 61.9 & 82.5 & 87.5 & 84.1 & 87.9 & 81.5\\	
		\bottomrule
	\end{tabular}
\end{table*}
To encode topological interaction between nodes into feature representations, GNNs generally integrate information from various neighborhoods via message passing. The consequence is that nodes in close proximity are more likely to be similar in feature representation. In other words, GNNs attempt to smooth out the difference between connected nodes, which is expected to favor the downstream classification tasks. However, in some settings where graphs have low edge homophily ratios, the vanilla GNNs have been found to underperform MLPs~\cite{zhu2020}. Here we tap into the impact of heterophilous edges on several popular GNN models by comparing the performance change of GNNs when heterophilous edges are perturbed. 

We first conduct an empirical study to test the node classification performance of vanilla GNNs in the ideal setting where the homophilious edges in the graph are retained, while all heterophilious edges are deleted. From the table~\ref{table:pilot} (see Section~\ref{sec:experiments} for the details of datasets and data split), we have the following observations:
\begin{itemize}
	\item When the graph becomes perfectly homophilious, the performance of the GNNs is greatly improved, compared to that on original structure. This observation suggests that a large number of heterophilious edges in graphs with weak homophily (i.e., strong heterophily) can be noisy and interfere in feature representation learning, since they allow the "messages" to pass between classes.
	
	\item Deeply stacking convolution layers (e.g. SGC with 50 convolution layers) does not result in drops in GNNs' performance, implying that the over-smoothing effect is only negative for inter-class messaging, but helpful for intra-class messaging.
\end{itemize}
\begin{figure}[htpb]
	\centering
	\subfigure[Training]{
		\begin{minipage}[t]{0.5\linewidth}
			\centering
			\includegraphics[width=2.4in]{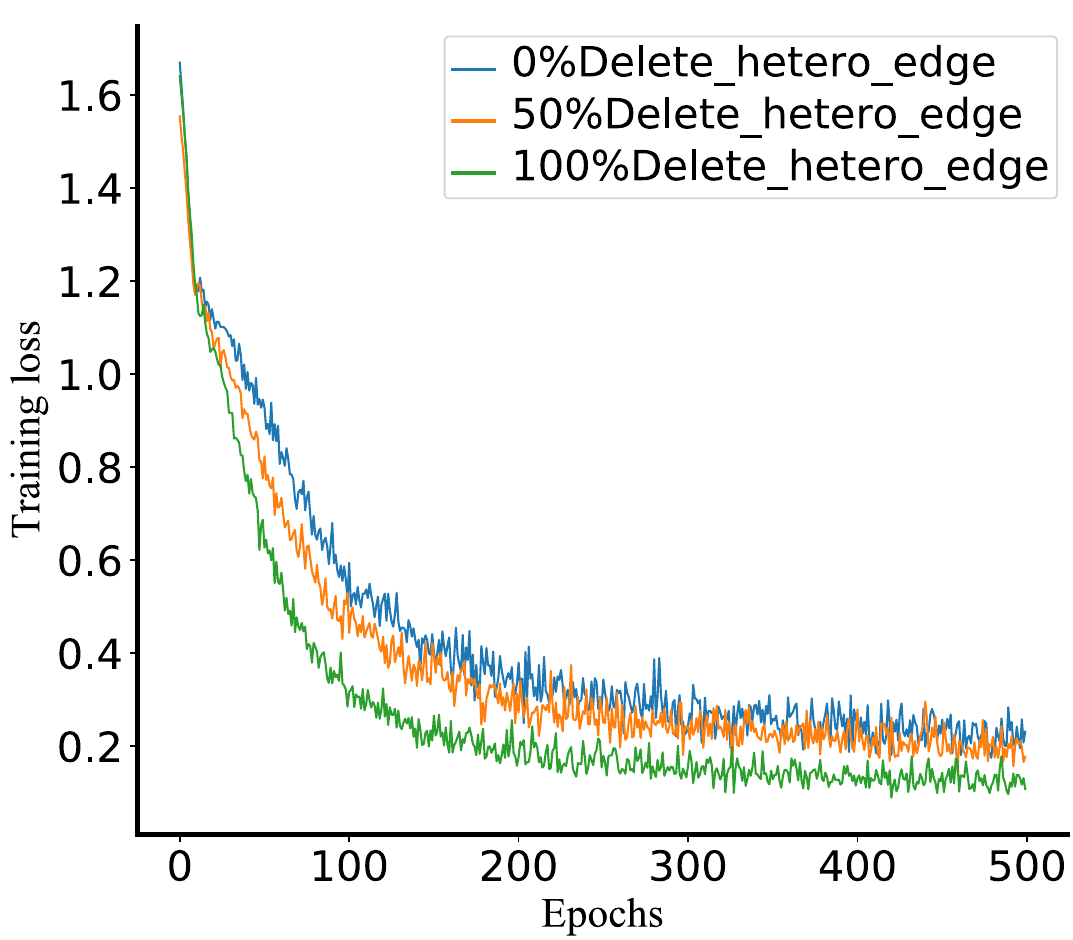}
		\end{minipage}%
	}\subfigure[Validation]{
		\begin{minipage}[t]{0.5\linewidth}
			\centering
			\includegraphics[width=2.4in]{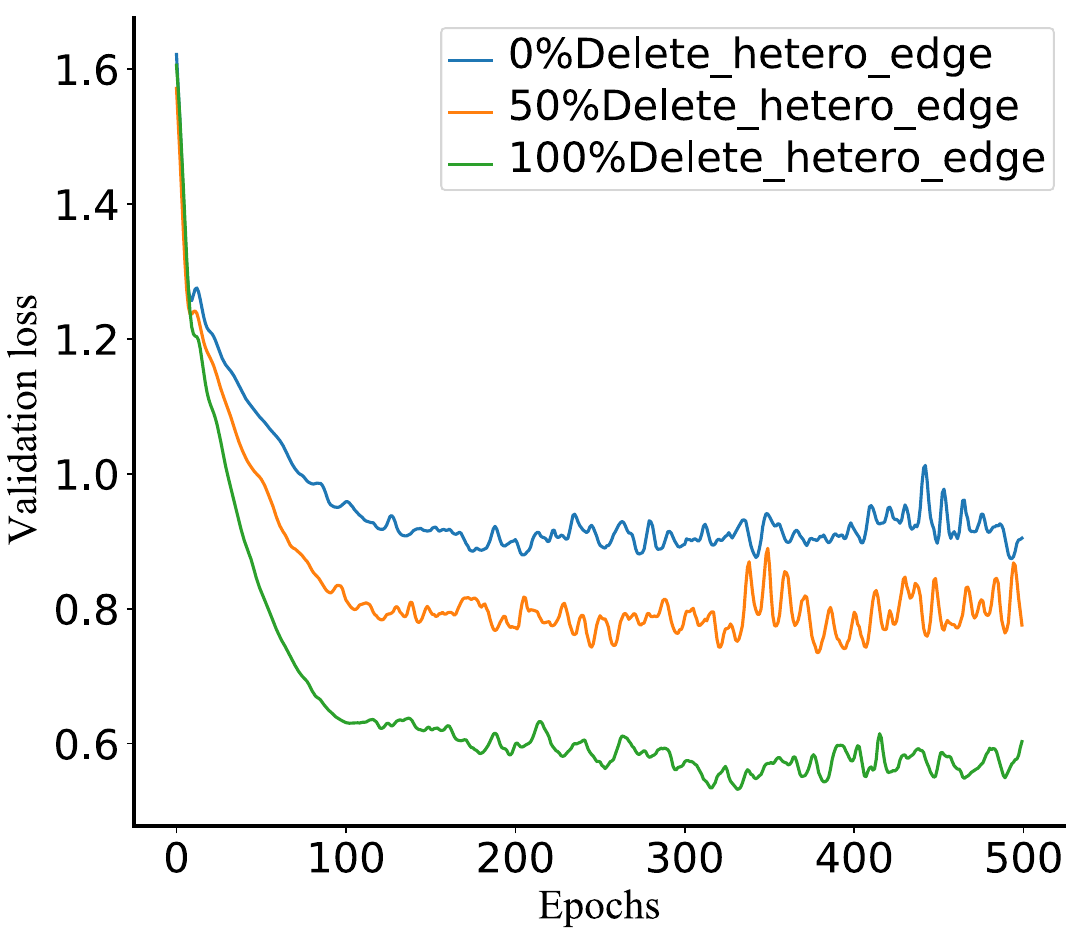}
		\end{minipage}%
	}
	\caption{Training and validation loss on Cornell under the heterophily edge deletion rates of 0\%, 50\%, and 100\% configurations.}
	\label{fig:pilot-loss}
\end{figure}

We further study the impact of heterophilious edges on graph learning via the training processes. Figure~\ref{fig:pilot-loss} shows the variations of training losses of SGC under the configurations where heterophilious edges are randomly removed with deletion rate of 0\%, 50\%, and 100\%, respectively. Note that 100\% deletion corresponds to the ideal situation. It is shown from the figure that the lowest loss of the model is achieved when there are no heterophilious edges, while the highest loss corresponds to the original graph structure, and deleting even a portion of heterophilious edges can help to better fit the training samples. The experimental results show the chances of exploiting heterophilious edges to make positive contribution to graph neural networks.

\section{Methodology}

We propose \emph{Spotting-then-Aggregating} (\emph{SA}), a novel scheme that aims to resolve the major limitation of message passing based GNNs when confronted with graph data that has strong heterophily. Instead of resort to high-order neighbors and estimating dissimilar neighbors (i.e., those connected with heterophilious edges), SA directly spots potential heterophilious edges, with which graph learning either performs on optimized structure (e.g., removing heterophilious edges from the graph such that message passing along those edges is not allowed), or integrates information from neighbors in a different way. To achieve this, we introduce an edge classifier that is pre-trained on training set at the first place, and then join it with a off-the-shelf or refined GNN model for further end-to-end training. The overall framework is diagrammatically illustrated in Figure~\ref{fig:framework}.

\subsection{Model Pretraining}
We first pre-train a binary classifier to initialize the graph update procedure.

\noindent\textbf{Pretraining task} We introduce a supervised edge classification task that predicts which type a given edge belongs to, where edges are divided into two categories, namely, homophilous and heterophilious edges. The edges among labeled nodes (i.e., the training set) are used for edge classifier pre-training, whose labels are obtained according to DEFINITION 1. In a similar way, the validation edge set and test edge set can also be obtained.

\noindent\textbf{Pre-training model} The representation of the edge between node $i$ and node $j$ is determined by the representations of the two endpoints:
\begin{equation}
	\bm{e}_{ij} = R(W\bm{x}_{i}, W\bm{x}_{j}, \gamma),
\end{equation}
where $\textbf{x}_{i}$ denotes the feature representation of node $i$, $W \in \mathbb{R}^{{f}'\times f}$ is a shared parameter matrix that maps node representations into a lower dimensional subspace, and $R(\cdot)$ can be any permutation invariance function. Here we consider two operations, namely, element-wise summation or squared difference, 

\begin{equation}
R(W\bm{x}_{i}, W\bm{x}_{j}) = (W\bm{x}_{i} - W\bm{x}_{j})^{2}, 
\end{equation}

Using edge representations as input features, a binary classifier is learned:
\begin{equation}\label{edge-c}
	\hat{y}_{ij} = softmax(W_{c}\boldsymbol{e}_{ij} + b_{c}),
\end{equation}
where $\hat{y}_{ij} \in \mathbb{R}^{2} $, and $W_{c},b_{c}$ are learnable parameters. For simplicity, here we employ a single layer feedforward network. However, we note that  any binary classifiers that are readily available can be used for edge classification. We use the cross-entropy loss to train the model. The prediction score can be further binarized to indicate the presence or absence of an edge in the changed graph, where the presence of an edge implies that it is identified to be homophilous edge. 

Next, we present two ways to exploit the inferred edge types in the follow-up graph representation learning.

\subsection{Method 1: Structure Optimization}
In this method, inspired by the above investigation on the impacts of heterophily, heterophilous edges identified by the edge classifier are removed to produce a homophilious graph, on top of which GNNs are expected to be improved. So the pre-trained edge classification model is joint with a GNN model, to update the parameters simultaneously. Here, we design an end-to-end training scheme. By Equation 6, we can get the kind of each edge, so we set the adjacency matrix as:
\begin{equation}
{A}^{c}_{i,j} =\phi( \hat{y}_{i, j})\quad s.t. A_{i,j}=1,
\end{equation}
where $\phi=1$ if the result of $\hat{y}_{i,j}$ is a homophilous edge, and 0 otherwise. We adopt SGC to learn graph structure. For SGC, 1-layer perceptrons can behave much like linear mappings, so the embedding expression ability of a layer of MLP is insufficient, especially when the original features of the node have a high dimensionality. Therefore, we use the SGC of two-layer MLPs as the base model and record it as SGC{\scriptsize 2}, it can be expressed as the following formula,
\begin{equation}
	\widehat{Y} = softmax({S^{e}_{sym}}^{K}XW_{0}W_{1}).
\end{equation} 
In this scheme, GNNs' downstream task is used to supervise the feature learning and structure update. Specifically, task-relevant training loss will be back propagated to the edge classifier and GNN at the same time.

However, to propagate the loss to the edge classifier is challenging. The reason is that as the output of edge classifier, the discrete graph structure is not differentiable. We tackle this problem by proposing a BackMax method inspired by the Gumbel-Softmax~\cite{gumbelsoftmax2017}.

\paragraph{BackMax} Our BackMax is detailed as follows.\\
1. For any arbitrary edge with endpoints \{$u, v$\}, the probability that the edge belongs to one of two classes is given by Equation.~\ref{edge-c}, i.e.,
\begin{equation}
	\boldsymbol{z}_{u,v} = edge\_classifier(\boldsymbol{x}_{u}, \boldsymbol{x}_{v}),
\end{equation}
where $\boldsymbol{z}_{u,v}\in \mathbb{R}^{2}$.\\
2. The softmax function is employed to amplify the probability distribution,
\begin{equation}
	\boldsymbol{\pi}_{u,v}=\frac{exp(\boldsymbol{z}_{u,v})}{\sum exp(\boldsymbol{z}_{u,v})},
\end{equation}
3. Then in the forward propagation phase, 
\begin{equation}
y_{hard} = one\_hot(argmax(\boldsymbol{\pi}_{u,v})),
\end{equation}
which is adopted to update the graph structure, while in the backward propagation phase, the error will be fed back to the continuous counterpart of $y_{hard}$: $y = \boldsymbol{\pi_{u,v}}$ for updating the parameters of the edge classifier. The following expression combines two variables together:
\begin{equation}
		\hat{y} = detach(y_{hard}-\boldsymbol{\pi}_{u,v})+\boldsymbol{\pi}_{u,v},
\end{equation}
where the function $detach()$ is to disconnect the backpropagation.  In comparison to Gumbel-softmax estimator~\cite{gumbelsoftmax2017}, BackMax does not involve any sampling for reparametrization and hyper-parameter for annealing, which can reduce the computational complexity.

\begin{thm}\label{thm1}
	SA($\cdot$) unifies MLP($\cdot$) and GNN($\cdot$) in supervised graph learning. 
\end{thm}
\noindent\emph{Proof.} Given a GNN {\bm$\hat{Y} = f(A,X,W)$}, in one extreme case where the edges among training nodes are all heterophilious, SA will learn to delete all connections in the graph, i.e.,  the adjacency matrix {\bm$A = I$} at the convergence. Then SA and MLP are equivalent (i.e.,  {\bm$\hat{Y} = f(A,X,W) = f(X, W)$}). On the other end of the spectrum, there are only homophilous edges in the training set, which makes the edge classifier bias to homophilous edges, so SA will not remove any links in the graph.  In this case, SA equals GNN.  When the training set is in between, SA will adaptively drop heterophilous edges that are assumed to be adverse to GNN's message passing. After structure optimization, for the nodes that are disconnected from the graph, SA behaves like MLP, while for the connected components, GNN will be performed.

\subsection{Method 2: Aggregation Optimization}
\label{sec:use_del_edge}
It can be found that structure optimization is based on the assumption that heterophilous edges are harmful to graph learning. In effect, the presence of heterophilous edges may convey information about the local connectivity of the nodes. Motivated by this, we revise the canonical message passing mechanism in the light of the type of edges associated with central node: 
\begin{equation}\label{agg-1}
    \bm{h}^{l+1}_{i} = \bm{h}^{l}_{i} + \sum_{j: \hat{y}_{ij}[1] = 1} \frac{1}{\sqrt{d^{c}_{i}}\sqrt{d^{c}_j}} \bm{h}^{l}_{j} W - \sum_{j: \hat{y}_{ij}[2] = 1}  \frac{1}{\sqrt{d^{e}_{i}}\sqrt{d^{e}_{j}}} \bm{h}^{l}_{j} W,
\end{equation}
where $\hat{y}_{ij}$ is the one-hot edge classification vector, whose first dimension indicates the homophilious edge between node $i$ and $j$ and second dimension indicates the heterophilous relationship between $i$ and $j$. $d^{c}_{i}$ denotes the number of homophilious neighbors of node $i$ and $d^{e}_{i}$ is the number of heterophilous neighbors of node $i$.  The second term in Eq.13 will force the central node $i$ approaching to the mean field of the neighborhood consisting of the nodes with the same label, that is, similar nodes will become more similar (i.e., smoothing effect). Meanwhile, the last term will distance the central node away from dissimilar neighbors (connected by heterophilous edges). So the messages passing through heterophilous edges can be deemed to diversify the representations of the nodes with different class labels. 

The new message passing mechanism has its explanation from the point of view of graph signal processing. It is well known that the canonical message passing based GNNs such as GCN~\cite{kipf2017gcn} and its variants is the approximation of the low-pass filtering on normalized symmetrix Laplacian:
\begin{equation}
\label{eq:low-pass}
\mathcal{F}_{L} =I-L = D^{-\frac{1}{2}}AD^{-\frac{1}{2}},
\end{equation}
So the signals on graph can be filtered with $\mathcal{F}_{L}$:
\begin{equation}
\label{eq:low-pass2}
(\mathcal{F}_{L \ast G})x = U(I - \Lambda) U^{T}x = \mathcal{F}_{L}\cdot x,
\end{equation}
 where the convolutional kernel in GCN is $g_{\theta} = (I - \Lambda)$. As the eigenvalues $\lambda_i$ of L range in [0, 2], the values in $g_{\theta}$ vary in [-1, 1], $g_{\theta}$ will encourage the signals at low frequency band (corresponding to small eigenvalues of $L$, i.e., $\lambda_i$ < 1). To also retain  signals at high frequency, an intuitive way is to make the convolutional kernel $g_{\theta}$ enlarge signals at the high frequency band, i.e., $g_{\theta} > 0$ for $\lambda_i > 1$. Among the possible filters, a simple one is $g_{\theta} = \Lambda$, whose parameters are in range [0, 2] so that the signals will be distilled by large eigenvalues corresponding to high frequency band [1, 2]. The high-frequency filtering is formulated as follows:
\begin{equation}
    (\mathcal{F}_{H \ast G})x = U\Lambda U^{T}x = \mathcal{F}_{H}\cdot x,
\end{equation}
\begin{equation}\label{hf}
    \mathcal{F}_{H} = L = I - D^{-\frac{1}{2}}AD^{-\frac{1}{2}}.
\end{equation}
 It is noteworthy that the elements of the filter $\mathcal{F}_{H}$ in Eq.~\ref{hf} are negative except for the diagonal elements that are positive, which exactly corresponds to the negative summation of the neighbors' information in the spatial domain. This property implies that high frequency channel can be used to learn the diversity among neighboring nodes. Accordingly, by combing high-frequency features with low-frequency ones graph representation learning is able to be enhanced.

Considering that the neighbors connected by heterophilous edges are different from central node in class labels, we adopt high frequency channel for heterophilous neighbors' message passing and low frequency channel for homophilous neighbors' message passing. Specifically,
we decompose the adjacency matrix into two subgraphs: the heterophilous subgraph consisting of inferred heterophilous edges $A^e$, and the homophilious subgraph that is the remaining by removing heterophilous edges from the original graph, i.e., ${A^{c}} = A - {A^{e}}$. Then the representation of each layer of non-parametric message passing can be written as:  
\begin{equation}\label{agg-2}
    H^{l+1} = S^{e}_{sym}H^{l} - \alpha S_{sym}^{c}H^{l},
\end{equation}
where $S^{e}_{sym}$ and $S_{sym}^{c}$ are the normalized symmetric adjacency matrices corresponding to homophilious subgraph $A^c$ and heterophilous graph $A^e$, respectively. $\alpha$ is a weighting parameter ranging in [0, 1], which can be predefined or learned. Clearly, Eq.~\ref{agg-2} is a general form of Eq.~\ref{agg-1}.


\section{Experiments}
\label{sec:experiments}
In this section, we evaluate the performance of SA on transductive node classification task on a wide variety of benchmark graph datasets. We use ''SA-base model'' and ''SA*-base model'' to represent the two implementations of the proposed framework with GNNs: structure optimization and aggregation optimization, respectively. 

\noindent\textbf{Datasets. } We conduct experiments on nine open graph datasets~\cite{2016citedatasets,wikidata2019,filmdata2019,pei2020} across the full spectrum of homophily ratio $h$. The statistics of the datasets is listed in Table~\ref{tb:dataintroduce}. For fair comparison, we follow the data partition in ~\cite{pei2020} (i.e., 48\%/32\%/20\% of nodes per class for train/validation/test) and adopt the shared 10 random splits for each dataset.
\begin{table}[htbp]
	\centering
	\caption{The statistics of the datasets}
	\label{tb:dataintroduce}
	\begin{tabular}{llllllllll}
		\toprule
		\textbf{Datasets} & \textbf{Nodes} & \textbf{Edges} & \textbf{Features} & \textbf{Classes} \\
		\midrule
		Cora & 2708 & 5429 & 1433 & 7 \\
		Citeseer & 3327 & 4732 & 3703 & 6 \\
		Pubmed & 19717 & 44338 & 500 & 3 \\
		Chameleon & 2277 & 36101 & 2325 & 5 \\
		Squirrel & 5201 & 217073 & 2089 & 5 \\
		Film & 7600 & 33544 & 931 & 5 \\
		Cornell & 183 & 295 & 1703 & 5 \\
		Texas & 183 & 309 & 1703 & 5 \\
		Wisconsin & 251 & 499 & 1703 & 5 \\
		\bottomrule
	\end{tabular}
\end{table}

\setlength\tabcolsep{3pt}
\begin{table}[htbp]
	\centering
	\caption{Test node classification accuracies (in percent) on graph datasets. Average accuracy and standard deviation over different splits is reported. ``\dag'' denotes results obtained from~\protect\cite{zhu2020}. Suffix (st) indicates that separate training is adopted in SA. The best results are in bold and the second best results are underlined.}
	\label{tb:result}
	\begin{tabular}{llllllllllllc}
		\toprule
		\textbf{Datasets} & \textbf{Cora} & \textbf{Cite.} & \textbf{Pubm.} & \textbf{Cham.} & \textbf{Squi.} & \textbf{Film} & \textbf{Corn.} & \textbf{Texa.}  & \textbf{Wisc.} &\multirow{2}{1cm}{ \textbf{Avg. Rank}}\\
		\textit{Homo.ratio} $h$ & 0.81 & 0.74 & 0.8 & 0.23 & 0.22 & 0.22 & 0.2 & 0.06 & 0.1  \\
		\midrule
		SGC & 83.72\scriptsize $\pm$1.4 & 74.01\scriptsize $\pm$1.7 & 83.62\scriptsize $\pm$0.5  & 47.41\scriptsize $\pm$2.5 & 36.68\scriptsize $\pm$1.4 & 25.49\scriptsize $\pm$1.5 & 59.45\scriptsize $\pm$5.2 & 58.37\scriptsize $\pm$4.2 & 51.56\scriptsize $\pm$8.8 & \makecell[c]{15}\\
		SGC{\scriptsize 2} & 86.90\scriptsize $\pm$1.6 & 76.31\scriptsize $\pm$1.4 & 87.03\scriptsize $\pm$0.7 & \underline{61.70\scriptsize $\pm$2.0} & 42.03\scriptsize $\pm$1.8 & 26.30\scriptsize $\pm$1.0 & 63.70\scriptsize $\pm$4.0 & 59.18\scriptsize $\pm$4.4 & 54.20\scriptsize $\pm$3.6 & \makecell[c]{11.3}\\
		GCN & 87.24\scriptsize $\pm$1.3 & 76.41\scriptsize $\pm$1.6 & 87.30\scriptsize $\pm$0.7 & 60.26\scriptsize $\pm$2.4 & 36.68\scriptsize $\pm$1.7 & 30.09\scriptsize $\pm$1.0 & 57.03\scriptsize $\pm$4.7 & 59.46\scriptsize $\pm$5.3 & 59.80\scriptsize $\pm$7.0 & \makecell[c]{11.4}\\
		GAT & 87.55\scriptsize $\pm$1.2 & 76.33\scriptsize $\pm$1.2 & 87.62\scriptsize $\pm$0.5 & 45.13\scriptsize $\pm$1.9 & 28.39\scriptsize $\pm$1.4 & 29.03\scriptsize $\pm$0.9 & 58.38\scriptsize $\pm$3.7 & 59.10\scriptsize $\pm$4.3 & 53.14\scriptsize $\pm$4.5 & \makecell[c]{14}\\
		MixHop\dag & 83.10\scriptsize $\pm$2.0 & 70.75\scriptsize $\pm$3.0 & 80.75\scriptsize $\pm$2.3 & 46.10\scriptsize $\pm$4.7 & 29.08\scriptsize $\pm$3.8 & 25.43\scriptsize $\pm$1.9 & 67.84\scriptsize $\pm$9.4 & 74.59\scriptsize $\pm$8.9 & 71.96\scriptsize $\pm$ 3.7 & \makecell[c]{15.7}\\
		GraphSAGE\dag & 86.60\scriptsize $\pm$1.8 & 75.61\scriptsize $\pm$1.6 & 88.01\scriptsize $\pm$0.8 & 58.71\scriptsize $\pm$2.3 & 41.05\scriptsize $\pm$1.1 & 34.37\scriptsize $\pm$1.3 & 75.59\scriptsize $\pm$5.2 & 82.70\scriptsize $\pm$5.9 & 81.76\scriptsize $\pm$5.6 & \makecell[c]{9.3}\\
		GCN-Cheby\dag & 86.68\scriptsize $\pm$1.0 & 76.25\scriptsize $\pm$1.8 & 88.08\scriptsize $\pm$0.5 & \underline{63.38\scriptsize $\pm$1.4} & 40.86\scriptsize $\pm$1.5 & 33.80\scriptsize $\pm$0.9 & 71.35\scriptsize $\pm$9.9 & 78.65\scriptsize $\pm$5.8 & 77.45\scriptsize $\pm$4.83 & \makecell[c]{9.1}\\
		\midrule
		GEOM-GCN\dag & 85.27 & \underline{77.99} & \textbf{90.05} & 60.90 & 38.14 & 31.63 & 60.18 & 67.57 & 64.12 & \makecell[c]{9.2}\\
		H2GCN-1 & 86.35\scriptsize $\pm$1.6 & 76.85\scriptsize $\pm$1.5 & 88.50\scriptsize $\pm$0.6 & 58.84\scriptsize $\pm$2.1 & 34.82\scriptsize $\pm$2.0 & \underline{35.94\scriptsize $\pm$1.3} & 79.45\scriptsize $\pm$6.7 & \underline{84.56\scriptsize $\pm$7.1} & 82.55\scriptsize $\pm$3.7 & \makecell[c]{6.4}\\
		H2GCN-2 &\underline{88.13\scriptsize $\pm$1.4} & 76.73\scriptsize $\pm$1.4 & 88.46\scriptsize $\pm$0.7 & 59.56\scriptsize $\pm$1.8 & 35.65\scriptsize $\pm$1.9 & 35.55\scriptsize $\pm$1.6 & 78.38\scriptsize $\pm$4.8 & \textbf{84.59\scriptsize $\pm$6.8} & 82.35\scriptsize $\pm$4.3 & \makecell[c]{5.9}\\
		FAGCN & 87.87\scriptsize $\pm$0.8 & 76.76\scriptsize $\pm$1.6 & 88.80\scriptsize $\pm$0.6 & 45.13\scriptsize $\pm$2.2 & 31.77\scriptsize $\pm$2.1 & 34.51\scriptsize $\pm$0.7 & 72.43\scriptsize $\pm$5.6 & 67.84\scriptsize $\pm$4.8 & 77.06\scriptsize $\pm$6.3 & \makecell[c]{10.0} \\
		CPGNN-MLP & 86.84\scriptsize $\pm$0.92 & 76.61\scriptsize $\pm$1.7 & 88.02\scriptsize $\pm$0.4 & 46.56\scriptsize $\pm$2.0 & 32.41\scriptsize $\pm$1.3 & 35.69\scriptsize $\pm$1.0 & 81.09\scriptsize $\pm$6.0 & 80.42\scriptsize $\pm$5.3 & 83.16\scriptsize $\pm$7.8 & \makecell[c]{7.6}\\
		CPGNN-Cheby & 85.27\scriptsize $\pm$0.73 & 75.56\scriptsize $\pm$2.3 & 87.14\scriptsize $\pm$0.2 & 46.34\scriptsize $\pm$3.1 & 29.58\scriptsize $\pm$2.3 & 35.10\scriptsize $\pm$1.2 & 80.54\scriptsize $\pm$6.0 & 80.27\scriptsize $\pm$4.3 & 83.14\scriptsize $\pm$3.3 & \makecell[c]{10.6}\\
		GPRGNN & 87.98\scriptsize $\pm$1.2 & 77.08\scriptsize $\pm$1.7 & 87.53\scriptsize $\pm$0.4 & 46.64\scriptsize $\pm$1.7 & 31.58\scriptsize $\pm$1.2 & 34.61\scriptsize $\pm$1.2 & 80.26\scriptsize $\pm$8.1 & 78.44\scriptsize $\pm$4.4 & 83.02\scriptsize $\pm$4.2 & \makecell[c]{8.3}\\
		GBK-GNN & \textbf{88.69\scriptsize $\pm$0.9} & \textbf{79.18\scriptsize $\pm$1.7} & \underline{89.11\scriptsize $\pm$2.3} & 61.62\scriptsize $\pm$1.7 & 41.70\scriptsize $\pm$1.0 & 35.66\scriptsize $\pm$1.2 & 78.81\scriptsize $\pm$4.8 & 80.67\scriptsize $\pm$4.4 & 80.00\scriptsize $\pm$3.8 & \makecell[c]{4.4}\\
		\midrule
		MLP & 75.13\scriptsize $\pm$2.7 & 73.26\scriptsize $\pm$1.7 & 85.69\scriptsize $\pm$0.3 & 46.93\scriptsize $\pm$1.7 & 29.95\scriptsize $\pm$1.6 & 34.78\scriptsize $\pm$1.2 & 79.79\scriptsize $\pm$4.2 & 79.19\scriptsize $\pm$6.3 & 83.15\scriptsize $\pm$5.7 & \makecell[c]{11.6}\\
		\midrule
		SA-SGC{\scriptsize 2} & 87.21\scriptsize $\pm$1.6 & 76.83\scriptsize $\pm$1.8 & 87.15\scriptsize $\pm$ 0.8 & 60.20\scriptsize $\pm$2.1 & \underline{44.21\scriptsize $\pm$2.2} & 34.96\scriptsize $\pm$1.0 & \underline{81.12\scriptsize $\pm$6.1} & 81.42\scriptsize $\pm$4.8 & \underline{
		84.52\scriptsize $\pm$4.0} & \makecell[c]{5.6}\\
		\textbf{SA*-SGC{\scriptsize 2}} & 87.36\scriptsize $\pm$1.2 & 76.88\scriptsize $\pm$1.7 & 87.55\scriptsize $\pm$ 0.6 & \textbf{64.94\scriptsize $\pm$2.5} & \textbf{48.97\scriptsize $\pm$1.1} & \textbf{38.28\scriptsize $\pm$2.2} & \textbf{81.12\scriptsize $\pm$5.8} & 83.52\scriptsize $\pm$5.8 & \textbf{
		84.52\scriptsize $\pm$4.0} & \textbf{\makecell[c]{3.1}}\\
		\bottomrule
	\end{tabular}
\end{table}

\noindent\textbf{Baseline Models.} We compare our method to strong baselines and state-of-the-art approaches, including GCN~\cite{kipf2017gcn}, SGC~\cite{2019sgc}, GAT~\cite{2018gat}, GCN-Cheby~\cite{cheyGCN2016}, graphSAGE~\cite{graphSAGE2017}, and Mixhop~\cite{mixhop2019}. We also compare our model with heterophily-oriented methods, namely, two variants of H2GCN (i.e., H2GCN-1 and H2GCN-2)~\cite{zhu2020}, Geom-GCN~\cite{pei2020} which quantitatively analyzes the homogeneity of a graph for the first time and the every recent work GPRGNN, CPGNN \cite{GPRGNN2021,CPGNN2021}. In particular, we choose the best two among the four variants of CPGNN for comparison. Moreover, we compare our model with recently proposed FAGCN~\cite{21FAGCN} and GBK-GNN~\cite{gbkgnn} that also leverage high-frequency signals. Specifically, FAGCN aims to relieve the over-smoothing effect on disassortative networks, because similar to heterophily, disassortativity accounts for the feature that nodes from different classes tend to connect with each other. In contrast to our two-channel method, GBK-GNN adaptively learns the dissimilarity between neighboring nodes to apply different filtering strategies (i.e., low-frequency or high-frequency filtering).

\noindent\textbf{Setup.} We implement models in Pytorch and use Adam optimizer for parameter updates. We set learning rate to $0.005$ for the pretraining module, dimension to $64$ for linear transformation  and use L2 regularizor with regularization factor of $ 0.0005$ on the weights of the linear layer. For the base models SGC,  SGC{\scriptsize 2} and GCN, we use two-layer MLPs as the classification model on all datasets, whose hidden units are $64$, learning rate is $0.01$, and weight decay is $0.0005$. Furthermore, dropout ratio of $0.6$ is applied to both layers. For scaling hyper-parameter $\alpha$, we search in the range [0, 1] spacing with 0.1.

\subsection{Results}
\label{sec:result}
The experimental results on real data are reported in Table~\ref{tb:result}. We observe that SA using SGC
as base model shows consistently strong performance across the full spectrum of homophily. Specifically, SA*-SGC{\scriptsize 2} that uses heterophilous edges for high-frequency message passing achieves state-of-the-art performance on average, and SA-SGC{\scriptsize 2} with structure optimization is comparative to the strong state-of-the-art method H2GCN. Note that graphSAGE and GCN-Cheby perform well on the datasets with high heterophily, demonstrating the benefits of seperate embedding of ego and its neighbors and learning higher order neighbors, which consists of design factors of H2GCN~\cite{zhu2020neu}. Compared to vanilla SGC, it is shown that our design for structure optimization is able to significantly improve the expressiveness of GNNs and thus boost the base model. Another observation on graph data with low homophily levels is that among several heterophily-oriented methods, namely, GEOM-GCN, two best H2GCN models, FAGCN, CPGNN, GPRGNN and ours, optimizing structure or message passing (ours) is more effective than integrating high-order features (prior work). We note that SGC{\scriptsize 2} with 2-layer MLPs remarkably outperforms the original SGC which uses one-layer linear discriminator. It should also be noted that if a graph has strong heterophily (e.g., Texas and Wisconsin), our structure optimization method will delete a large proportion of the edges in the graph. In this case, the learned SGC model is close but still better than MLP.

It is noteworthy that compared to the more recent GBK-GNN that adaptively aggregates different types of features from neighbors, though our aggregation optimization method is slightly inferior to GBK-GNN on graphs with high homophily levels, it shows significant superiority over GBK-GNN on graphs with strong heterophily. The reason is that our aggregation optimization method offers the explicit information about the similarity between central node and its neighbors in terms of class labels such that features of different neighbors are discriminatively treated: homophilious neighbors are low-pass filtered while heterophilous neighbors are high-pass filtered. Meanwhile, it is much easier for edge classifier to spot heterophilous edges in graphs with strong heterophily than in graphs with high homophily, which will be demonstrated in detail in the following subsection. As a result, aggregation optimization brings more gains for graphs with strong heterophily.   

\subsection{Performance of Edge Classification}
\label{sec:deletedge}
Edge classification is critical to the effectiveness of our framework. To look into the relationship between edge classification and model performance on downstream task (i.e., node classification), we measure to what degree the edges are correctly classified after the model converges, i.e., the accuracy of edge type identification. Moreover, to validate whether SA can effectively screen out heterophilious edges, we consider the ratio of true heterophilious edge to the total of identified heterophilous edges, termed heterophilious edge recall.

\begin{table}[htbp]
	\centering
	\caption{Accuracy of edge classification.}
	\label{tb:edge_acc}
	\begin{tabular}{llllllllllllc}
		\toprule
		\textbf{Metrics} & \textbf{Cora} & \textbf{Cite.} & \textbf{Pubm.} & \textbf{Cham.} & \textbf{Squi.} & \textbf{Film} & \textbf{Corn.} & \textbf{Texa.}  & \textbf{Wisc.} \\
		\textit{Homo.ratio} $h$ & 0.81 & 0.74 & 0.8 & 0.23 & 0.22 & 0.22 & 0.2 & 0.06 & 0.1  \\
		\midrule
		Accuracy & 76.89\scriptsize $\pm$1.2 & 77.88\scriptsize $\pm$1.1 & 70.15\scriptsize $\pm$ 0.6 & 74.02\scriptsize $\pm$1.2 & 77.11\scriptsize $\pm$1.5 & 85.41\scriptsize $\pm$2.0 & 85.79\scriptsize $\pm$1.8 & 97.52\scriptsize $\pm$5.8 &
		85.92\scriptsize $\pm$2.0 \\
		
		Recall & 24.56\scriptsize $\pm$1.1 & 3.65\scriptsize $\pm$0.1 &	20.30\scriptsize $\pm$0.8 &	25.65\scriptsize $\pm$2.2  & 26.00\scriptsize $\pm$2.4 &	94.2\scriptsize $\pm$1.8 & 100.00\scriptsize $\pm$0.0 & 100.00\scriptsize $\pm$0.0 & 100.00\scriptsize $\pm$0.0\\

		\bottomrule
	\end{tabular}
\end{table}


The results in Table~\ref{tb:edge_acc} suggest that, the homophily level has an impact on edge classification bias, that is, the edge type classifier performs better on graphs with low homophily level than on highly homophilious graphs. More precisely, in graphs that have strong homophily/heterophily, homophilous/heterophilious edges are far more than heterophilious/homophilous edges. This imbalance is favorable when it is biased towards heterophilious edges. On the other hand, it will be undesirable for structure optimization when homophilous edges dominate, as in this setting the classifier has a higher chance to mistakenly infer an edge to be homophilous, whose ground-truth actually is heterophilious. As the recall index in the table suggests, heterophilous edges are successfully spotted in graphs with strong heterophily, while their identification suffers from homophily bias.  

Recall that method 2 involves two channels of message passing: positive aggregation for homophilious neighbors (i.e., neighbors with the label same as central node's), and negative aggregation for heterophilous neighbors (i.e., neighbors with the labels different from central node's). So high-accuracy edge classification favors the positive message passing between intra-class nodes and negative message passing between inter-class nodes, which  explains why SA performs better on graphs with low homophily.



\subsection{Comparing Method 1 to DropEdge}
\label{sec:dropedge}
Our structure change strategy (i.e., method 1) has connection to the previous method DropEdge~\cite{dropedge20}, as both of them are to remove links in graphs. However, different from DropEdge~\cite{dropedge20} that randomly delete some links, structure optimization aims at removing some heterophilous edges that are optimal-related to performance. In this section, we compare our method with DropEdge to verify the effectiveness of our method. Note that DropEdge randomly removes a portion of edges in the graph during training, so for fair comparison, we set the drop ratio in DropEdge to be equal to the total edge deletion ratio obtained by our method on each dataset.

\begin{table}[htbp]
	\centering
	\caption{Test classification accuracies (in percent)}
	\label{tb:vsdopedge}
	\begin{tabular}{llllllllll}
		\toprule
		\textbf{Drop rate} & \textbf{6\%} & \textbf{17\%} & \textbf{92\%} & \textbf{98\%} \\ 
		\textbf{Datasets} & \textbf{Cora} & \textbf{Squi}. & \textbf{Film} & \textbf{Texas}  \\
		\textbf{Edge} & \textbf{5429} & \textbf{217073} & \textbf{33544} & \textbf{309} \\
		\midrule
		SGC{\scriptsize 2} & 86.9\scriptsize $\pm$1.6 & \underline{42.4\scriptsize $\pm$1.5} & 25.3\scriptsize $\pm$1.0 & 59.2\scriptsize $\pm$4.4 \\
		GCN & 87.2\scriptsize $\pm$1.3 & 36.7\scriptsize $\pm$1.7 & 30.1\scriptsize $\pm$1.0 & 59.5\scriptsize $\pm$1.6 \\
		GAT & 87.55\scriptsize $\pm$1.2 & 28.39\scriptsize $\pm$1.4 & 29.03\scriptsize $\pm$0.9 & 59.10\scriptsize $\pm$4.3 \\
		\midrule
		Dropedge-SGC{\scriptsize 2} & 85.5\scriptsize $\pm$1.6$\downarrow$ & 39.0\scriptsize $\pm$2.0$\downarrow$ & 30.9\scriptsize $\pm$0.7 & 65.7\scriptsize $\pm$8.6 \\
		Dropedge-GCN & 84.9\scriptsize $\pm$1.3$\downarrow$ & 38.8\scriptsize $\pm$1.8 & 27.4\scriptsize $\pm$1.1$\downarrow$ & 58.9\scriptsize $\pm$5.0$\downarrow$ \\
		Dropedge-GAT & 82.11\scriptsize $\pm$1.4$\downarrow$ & 22.35\scriptsize $\pm$1.8$\downarrow$ & 21.06\scriptsize $\pm$4.1$\downarrow$ & 58.92\scriptsize $\pm$4.3$\downarrow$\\
		\midrule
		SA-SGC{\scriptsize 2} & 87.2\scriptsize $\pm$1.6 & \textbf{44.2\scriptsize $\pm$1.6} & \textbf{35.2\scriptsize $\pm$1.0} & \textbf{81.2\scriptsize $\pm$4.8} \\
		SA-GCN & \underline{87.4\scriptsize $\pm$1.2} & 41.9\scriptsize $\pm$1.7 & \underline{35.0\scriptsize $\pm$1.0} & \underline{80.0\scriptsize $\pm$5.2} \\
		SA-GAT & \textbf{87.6\scriptsize $\pm$1.1} & 40.51\scriptsize $\pm$2.1 & 31.04\scriptsize $\pm$5.5 & 79.43\scriptsize $\pm$3.7 \\
		\bottomrule
	\end{tabular}
\end{table}

Table~\ref{tb:vsdopedge} shows the change of accuracy when we apply two methods of dropping edge on two base models (SGC{\scriptsize{2}}, GCN and GAT). We can see that our goal-directed method is evidently superior to random deletion, while random deletion (i.e., DropEdge) may even degrade the performance of base models, especially GAT. Another advantage of our method compared to DropEdge is that SA-GNNs does not introduce any additional hyperparameters, e.g., edge deletion ratio, the model learns target edges to drop adaptively.

\subsection{Hyper-parameter in Method 2}
\label{sec:ablation}

\makeatletter\def\@captype{figure}\makeatother
\begin{minipage}{0.4\textwidth}\label{alpha}
\centering
\caption{Effect of hyper-parameter $alpha$ on node classification accuracy. }
\includegraphics[scale=0.20]{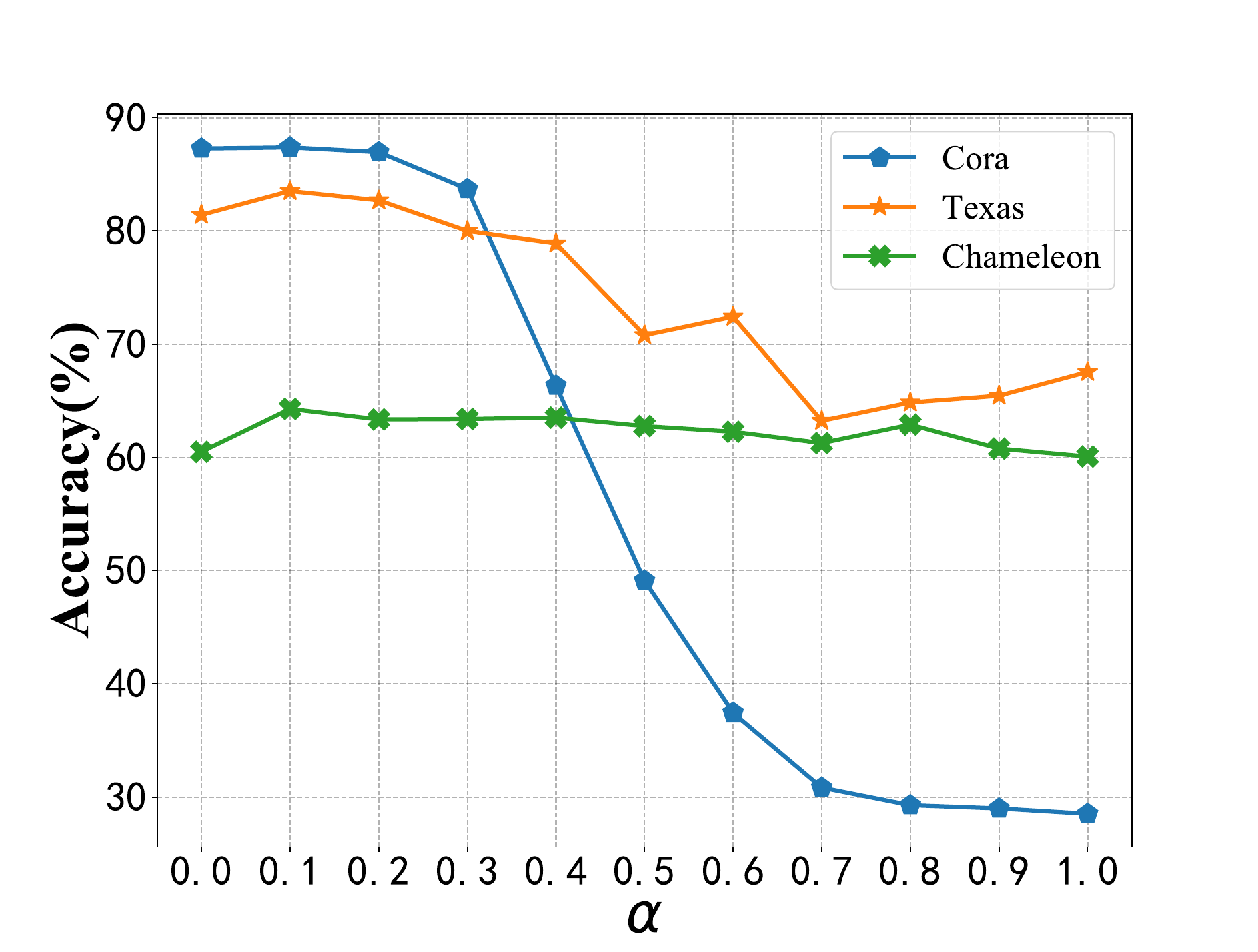}
\label{fig:alpha}
\end{minipage}
\makeatletter\def\@captype{table}\makeatother
\begin{minipage}{.7\textwidth}
\centering
\caption{Comparison of manual setting and adaptive learning of $alpha$.}
	\begin{tabular}{llllllllll}
	
		\toprule
		\textbf{Dataset} & $\alpha (learning)$ & $\alpha (manual)$ \\
		\midrule
		\textbf{Cora} & 86.82\scriptsize $\pm$1.2 & \textbf{87.36\scriptsize $\pm$1.2} \\
		\textbf{Cham.} & 59.65\scriptsize $\pm$2.0 & \textbf{64.94\scriptsize $\pm$2.2}\\
		\textbf{Texas} & 78.23\scriptsize $\pm$1.0 & \textbf{83.52\scriptsize $\pm$5.8}\\
		\bottomrule
		\label{tb:sel_alpha}
	\end{tabular}
\end{minipage}

In method 2 (i.e., aggregation optimization), hyper-parameter $\alpha$ weighs the influence of the features from heterophilous neighbors on node representation. It is interesting to explore the impact of $\alpha$ on the final performance. Results in Figure~\ref{fig:alpha} demonstrate that a high weighting on heterophilous features might hurt performance. In particular, for Cora that shows strong homophily, a large proportion of features  from heterophilous neighbors remarkably impairs the performance of GNNs, while for Texas and Chamleon with lower homophily levels, they are more robust to the combination of high-frequency features. In general, a small $\alpha$ around 0.1 is appropriate for graphs. 

Besides manual tuning, we also implement a trainable parameterized aggregation optimization, which allows $\alpha$ to be learned. Table~\ref{tb:sel_alpha} reports the performance of method 2 with two kinds of parameter searching techniques on three benchmark datasets. The results show that empirical setting of $\alpha$ is better than adaptive learning.

\section{Related Work}

\textbf{Homophily and Heterophily.} Homophily describes the consistency between node class labels and the graph structure. Prior work has shown the homophily plays a crucial role in graph learning. GeomGNNs~\cite{pei2020} first pays attention to this property and provides a metric to measure the homophily level of a graph. In particular, Geom-GCN leverages non-Euclidean embedding technique to better capture structural information and long-distance dependence. In recent work, Kim et al.~\cite{kim2020find} find that homophily can affect the power of attention mechanism. Recent work by Hou et al.~\cite{CSGNN2020} synthetically study the ability of GNN to capture graph information, and also propose metrics to measure feature smoothing and label smoothing quantitatively. To allow the model to be aware of the information not only about node features but about class labels, graph markov neural networks~\cite{2019gmnn} models the joint label distribution with conditional random field, which can be effectively trained with the variational EM algorithm. On the other hand, to mitigate the negative effect of heterophily on performance, H2GCN~\cite{zhu2020} studyies the distribution of heterophilious edges in graph and finds that on a graph with a high level of heterophily, the 2-hop neighborhood is always dominated by homophily nodes, so it can be used for feature representation learning. GPRGNN~\cite{GPRGNN2021} adaptively learns generalized pagerank weights so as to jointly optimize node feature and topological information extraction. In addition, CPGNN~\cite{CPGNN2021} shares similar motivations to ours, i.e., both works attempt to identify connected label pairs that are incompatible/compatible. But unlike CPGNN that learns compatibility for all label pairs via belief diffusion, our method directly learns an incompatible/compatible classifier connections. A recent work ~FAGCN~\cite{21FAGCN} can be regarded as another revelant work. FAGCN aims at disassortative data, in which nodes belonging to different communities (or classes) are connect to each other. To enrich the feature representation, FAGCN utilizes the differences between nodes (the so-called high frequency part of signal). HOG-GCN measures the homophily between node pairs through topology and feature information, and designs GNN that can automatically change the promotion and aggregation process. GBK GNN~\cite{gbkgnn} proposes a GNN
model based on a bi-kernel feature transformation and a selection
gate. This method shares some similarity with ours, but GBK-GNN is to adaptively learn different kernels, and it aggregates different types of edges by adding, while we explicitly distance heterophilous neighbors by subtracting the messages from those nodes. LINKX~\cite{LINKX} is the first method proposed in large-scale heterophilious graph datasets. It can achieve better results on large-scale heterophilious datasets simply by sending the feature matrix and adjacency matrix into MLP for learning and fusion. Generally, previous work mainly focuses on enhancing the expressivity of feature representations.  To the best of our knowledge, this work is the first attempt to resolve the heterophily challenge from the angle of edge type identification.

\noindent\textbf{Graph Sparsification.} The operation of discarding heterophilious edges in our work can be viewed as a way of graph sparsification. A popular graph sparsification model is graphSAGE~\cite{graphSAGE2017}, whose goal is to allow graph convolution to be available in large-scale graphs. Likewise, Dropedge~\cite{dropedge20} randomly drops edges to overcome the oversmoothness of graph networks. However,
most of the existing work on graph sparsification ignores the relatedness between edges and tasks, except for  NeuralSparse~\cite{neuralspace2020} that learns k-neighbor subgraphs for deleting task-irrelevant edges. In contrast, our model performs task-oriented sparsification.

\section{Conclusion and Discussion}
We have proposed a general framework to leverage the heterophily of graphs for boosting GNNs on graph data with low homophily levels from the perspective of edge type. On the conjecture that heterophilious edges mislead GNNs to aggregate information from different classes, we have first devised an optimal structure learning method based on edge type classification. Also based on edge classification, we have proposed to not optimize the graph structure but optimize the way of feature aggregation. Towards this end, we have introduced additional message passing channel to convey information about heterophilous neighbors. Our experiments on a variety of benchmark datasets show the significantly strong performance of the two proposed methods on graphs with high heterophily. 

However, it should be noted that since the edge classification will be biased towards the majority of a certain edge type, it remains an open challenge for improving the accuracy of edge classification on graphs with strong homophily, which we leave as future work.

\section*{Acknowledgments}
 {
	This work is partially supported by National Natural Science Foundation of China (No.61873218) and SWPU Innovation Base No.642.
}

\bibliographystyle{ACM-Reference-Format}
\bibliography{reference}

\appendix

\end{document}